\documentclass{article}
\usepackage{spconf,amsmath,amsfonts,epsfig}

\title{Automatic Text Area Segmentation in Natural Images}

\name{Syed Ali Raza Jafri, Mireille Boutin, and Edward J. Delp \thanks{This work was supported by a grant from Next Wave Systems, LLC.}}
\address{School of Electrical and Computer Engineering\\
Purdue University\\
465 Northwestern Av. \\
West Lafayette, IN 47907}
\begin{document}
%
\maketitle
%
\begin{abstract}
We present a hierarchical method for segmenting text areas in natural images.
The method assumes that the text is written with a contrasting color on a more or less uniform background.
But no assumption is made regarding the language or character set used to write the text.
 In particular, the text can contain simple graphics or symbols.
The key feature of our approach is that we first concentrate on finding the background of the text, 
before testing whether  there is actually text on the background.
Since uniform areas are easy to find in natural images, 
and since text backgrounds define areas which contain "holes" (where the text is written)
we thus look for uniform areas containing "holes" and label them as text backgrounds candidates.
Each candidate area is then further tested for the presence of text within its convex hull. 
We tested our method on a database of 65 images including English and Urdu text.
The method correctly segmented all the text areas in 63 of these images, 
and in only 4 of these were areas that do not contain text also segmented.
\end{abstract}
%
%

\section{Introduction}
\label{sec:intro}
Finding areas of text in a natural image is a difficult problem. 
One reason for this is that, from a texture and geometric standpoint, 
many objects  (e.g., tree branches or electrical wire on a blue sky) resemble text. 
In addition, the text size, font, color, orientation and skew are generally unpredictable. 
In the following, we introduce an efficient method for finding text areas in natural images.
The only assumption we make regarding the text is that it is written 
on a more or less uniform background
using a contrasting color.
In particular, the text could be written in cursive letters or with foreign characters and even contain simple graphics.
We proceed by using a combination of texture and shape features
in order to sequentially rule out regions of the image which do not contain text.
One of key elements of our approach
is that we begin by focusing on finding the text background, rather than the text itself. 
Finding backgrounds is a lot simpler than finding text directly.
It can be accomplished robustly by extracting some well chosen texture features.
Once a potential background area has been selected, 
we then use a combination of shape and color features to detect whether text is present inside the area. 
Having pre-identified the background provides us with a sample of the background color and texture, 
and thus simplifies the problem of determining whether there is text on the background.
The search for the text area is performed hierarchically in a top-down fashion:
if no text is found at a given scale, then we look for text at a smaller scale.
This allows us to find the text without making prior assumptions regarding the font and area sizes.

There is an extensive literature on the problem of segmenting text areas in images.
Many methods are specifically designed to deal with special types of images, such as documents  \cite{Amamoto93}, webpages \cite{Park98}, or pictures of newspaper  \cite{Yuan01}.
For each type of documents, assumptions are made regarding the type of text present in the image and its surrounding.
For example,  for newspapers it can be assumed that the text is written in black on a white background,
for text written using the roman alphabet it can be assumed that the characters are written at regular intervals, etc.
One problem that resembles the problem of finding text in natural area is that of segmenting license plates on natural images containing cars \cite{Cano03, Shapiro04},
on which a lot of effort has been put.

The general problem of finding text in natural images has also been considered \cite{Zhong95, Garcia00, Pietkainen01}.
Many text area segmentation methods formulate the problem as a two-class decision problem:
 is the given block/pixel part of a text area or not? Then the problem is reduced to finding good features for formulating this decision problem 
 and finding a good classifier for making decisions. 
 In general, three types of features are used for identifying the text: texture based, edge based or connected-component based. 
Sometimes, three or more classes are considered (e.g., text, background, and other) in order to increase the accuracy of the classification. 
But false positives always constitute a nuisance and the parameters of the classifier must be carefully tuned in order to avoid them as much as possible.
 In contrast, we do not look text directly, but rather find the text by first finding likely text contexts 
 and studying the features of each  potential text context to decide whether or not it contains text. 
 False positives in the early stages thus do not constitute a problem, 
 and so we can conservatively estimate the thresholds of the early decision parameters.  
 The details of our approach are given in the next section.
  In Section \ref{sec:experiments}, 
  we present our experimental methodology and results 
  before concluding in Section \ref{sec:conclusion}.

\section{Proposed Segmentation Method}
\label{sec:segmentation}

All the text areas that can be found in natural images have little in common.
They can be written in different languages (e.g., English, Korean, Arabic) with different fonts, using different colors, and on different backgrounds.
Shadows and occlusions can also be present.
The image acquisition process adds further variability such as  
camera skew,  distance, lighting and motion blur.
But almost all the text that is easily noticeable by humans 
has an important characteristic: 
it is written on a background that is more or less uniform, at least piecewise.
Another common characteristic is that the text color typically contrasts with the background color.

\begin{figure}[htb]
\begin{center}
 \includegraphics[height=12.0cm]{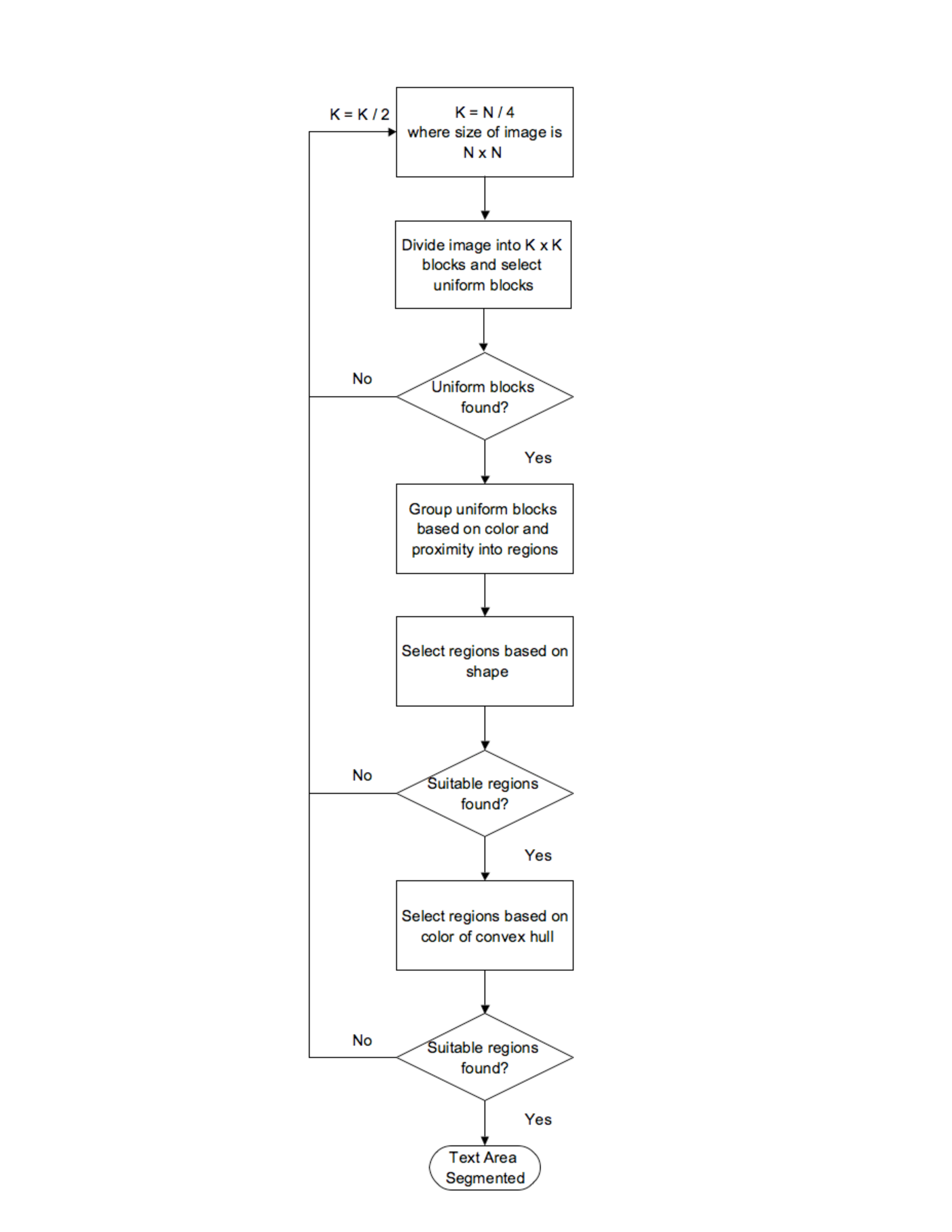}
\end{center}
\caption{Schematic Representation of the Steps of the Proposed Area Segmentation Method.}
\label{fig:steps}
\end{figure}

In order to look for text, we thus begin by looking for image areas where the color is more or less uniform. 
So we divide the image into blocks of equal size and test each block for uniformity. The initial block size is set to one quarter the image size. 
Once uniform blocks have been identified, we group them based on color similarity and proximity.
Groups whose shape is incompatible with that of a text background are then ruled out.
The interior of the convex hull of each group that remains
is then tested for the presence of text.
If no text is found, then a smaller block size is chosen and the above steps are repeated.
A schematic representation of this process is given in Figure \ref{fig:steps}.
We now describe each step in more details.

\subsection{Uniform Block Detection}
\label{sec:uniformity}

Once the image is divided into blocks of size $k\times k$, 
we select the blocks that have a uniform texture.  
One way to do this consists in looking 
at the color distribution of the pixels contained in the block,
and in checking to what extent this 3-D distribution is scattered: 
a low amount of scattering is an indication of a  uniform color.
However, this would not account for the fact that some backgrounds
are made of pixels of different colors 
that are arranged in such a way that they do not
form any visible higher level structures.
Such backgrounds do look uniform, 
even though their color distribution may be very scattered.

We therefore use an alternative approach 
to quantify the uniformity of a given block.
We consider each $k\times k$ color block 
as a sum of three one-color blocks: a red one, a green one and a blue one.
More precisely, after concatenating the rows of pixels of the block, we can represent it as a $k^2 \times 3$ matrix
\[
\left( 
\begin{array}{c}
{\bf p}_1\\
{\bf p}_2\\
\vdots \\
{\bf p}_{k^2}
\end{array}
\right) = 
\left( 
\begin{array}{ccc}
r_1  & g_1 & b_1\\
r_2 & g_2 & b_2 \\
\vdots  & \vdots & \vdots \\
r_{k^2} & g_{k^2} & b_{k^2}
\end{array}
\right).
\]

\begin{figure}[htb]
\begin{center}
 \includegraphics[height=7.0cm,width=7.0cm]{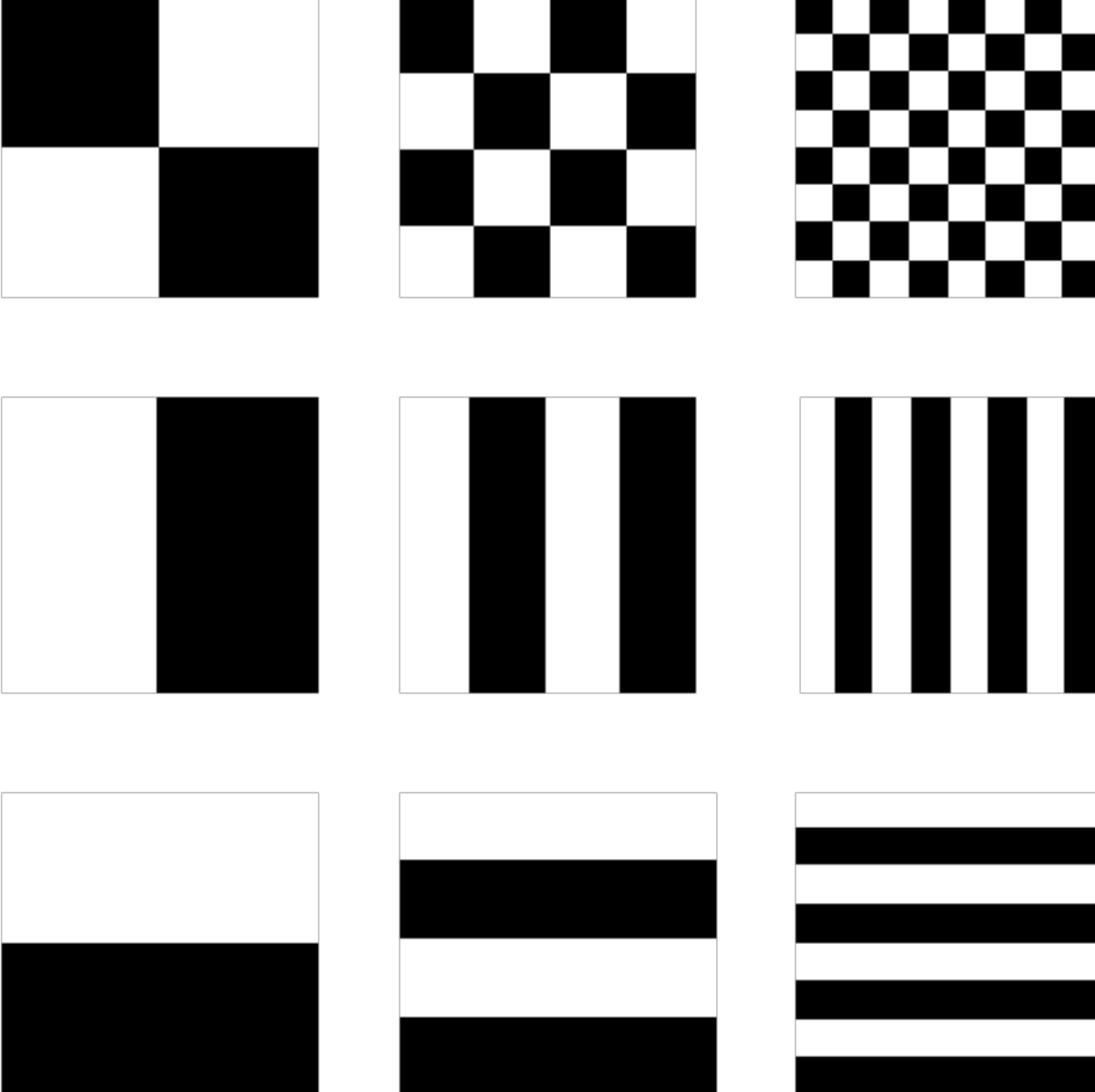}
\end{center}
\caption{Some Basis Vectors used as Filters to Quantify the Uniformity of the Blocks. The colors black and white are used to represent $+1$ and $-1$ entries respectively.}
\label{fig:basis}
\end{figure}

Then each row of the above matric (which corresponds to a $k\times k$ block of a single color) is projected onto a basis $\{ {\bf v}_1,{\bf v}_2,\ldots, {\bf v}_{k^2}\} $ 
of ${\mathbb R}^{k^2}$ which satisfies the following conditions:
\begin{enumerate}
\item the entries of the basis vectors consist of $+1$'s and $-1$'s
\item the sum of the entry of any basis element is equal to zero.
\end{enumerate}
In other words, 
the value of the projection of a row of the matrix onto any one of these basis vectors 
quantifies the difference between the amount of color on two regions of equal size within the block. 
Some elements of such a basis are illustrated in Figure \ref{fig:basis}.
The basis elements can be viewed as filters which quantify uniformity of the color of the block:
the smaller the projection of the three color blocks onto these basis elements,  the more uniform the texture of the given color block.

We can write 
\begin{multline*}
\left( 
\begin{array}{c}
r_1  \\
r_2 \\
\vdots   \\
r_{k^2} 
\end{array}
\right)
= \sum_{i=1}^{k^2} \alpha_i  {\bf v}_i,
\hspace{0.5cm}
\left( 
\begin{array}{c}
g_1  \\
g_2 \\
\vdots   \\
g_{k^2} 
\end{array}
\right)
= \sum_{i=1}^{k^2} \beta_i  {\bf v}_i ,\\
 \left( 
\begin{array}{c}
b_1  \\
b_2 \\
\vdots   \\
b_{k^2} 
\end{array}
\right)
= \sum_{i=1}^{k^2} \gamma_i  {\bf v}_i, \hspace{2cm}
\end{multline*}
where $\alpha_i$ is equal to the dot product between $(r_1,\ldots,r_{k^2})^T$ and the basis vector ${\bf v}_i$ divided by the norm of ${\bf v_i}$, 
and similarly for $\beta_i$ end $\gamma_i$.
 Observe that a block has a completely uniform color if and only if all the coefficients $\alpha_i$'s, $\beta_i$'s and $\gamma_i$'s are zero.

In our method, we take each block of the image and obtain the coefficients   $\alpha_i$,   $\beta_i$  and $\gamma_i$ of the expansion for each color of the block. We then choose the blocks for which the $L_2$ norm of these coefficients is less than one standard deviation below the mean of the other blocks inside the picture. This threshold was chosen empirically. For better performance, it could be learned from a training set.

\subsection{Uniform Block Grouping}
\label{sec:grouping}

Once the uniform blocks have been selected, we group them together in order to form uniform regions.
We begin by grouping sets of connected blocks based on color similarity. 
More precisely, we group together connected uniform blocks 
if the distance between their mean color vector is less than 45.
Again, this threshold value was chosen empirically. A better value could be obtained from a training set.
Once we have obtained connected uniform regions, 
we merge these regions based on color similarity 
and based on the variation of color in the space between them.
More precisely,  we merge regions such that 
\begin{enumerate}
\item the distance between their mean color vector is less than 45;
\item  the distribution of colors between the two regions is either unimodal or bimodal. If it is bimodal we look for at least a distance of 100 between the two peaks.
\end{enumerate}

\subsection{Shape Test for Groups of Uniform Blocks}
\label{sec:shape}

Since the image areas containing the text itself are not uniform, then any
uniform region corresponding to the background of a sign must have "holes". 
In other words,  we assume that the text is at least partially surrounded by a uniform area. 
Any selected uniform area which is connected and convex is thus eliminated.
This simple step rules out most of the uniform regions identified with the 
previous steps. The few remaining regions (if any) go through the next and final 
step of our method.  Note that one often needs to reach a small scale before 
a uniform region with an appropriate shape is identified.

\subsection{Test for Text Inside Convex Hull of Group}
\label{sec:text inside}

Given a uniform area with holes, we obtain the convex hull of that area and test whether text is present inside.
There are many sophisticated ways to test for the presence of text. But many of these ways are language specific.
Fortunately, we found that considering the color distribution of the convex  hull of the uniform region 
was enough to discriminate between text and non-text in most circumstances. 
Our test is based on the fact that text areas contain few and contrasting colors.
We thus look for color distribution which are multimodal and with a large distance between the modes.
More precisely, we expand our uniform regions using smaller block sizes until we
cover as much of the background as possible. We then determine the average color intensity of the
set of points within the convex hull that are not part of the background. The color distance between 
background and non-background points provides an accurate feature to determine which regions contain text.
Currently we use a color distance of 100 as a conservative minimum distance between text and background. 
Again, training could be used to select a better threshold.

\begin{figure*}[t]

\begin{minipage}[b]{0.24\linewidth}
  \centering
 \centerline{\epsfig{figure=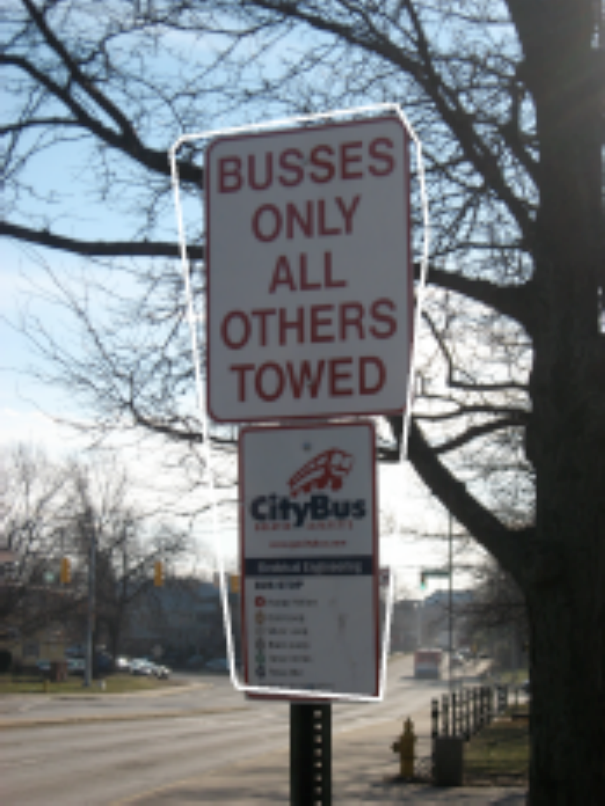,width=4 cm}}
  \centerline{(a)}\medskip
 \end{minipage}
  \hfill
  \begin{minipage}[b]{0.75\linewidth}
  
  \quad 
\begin{minipage}[b]{0.24\linewidth}
  \centering
 \centerline{\epsfig{figure=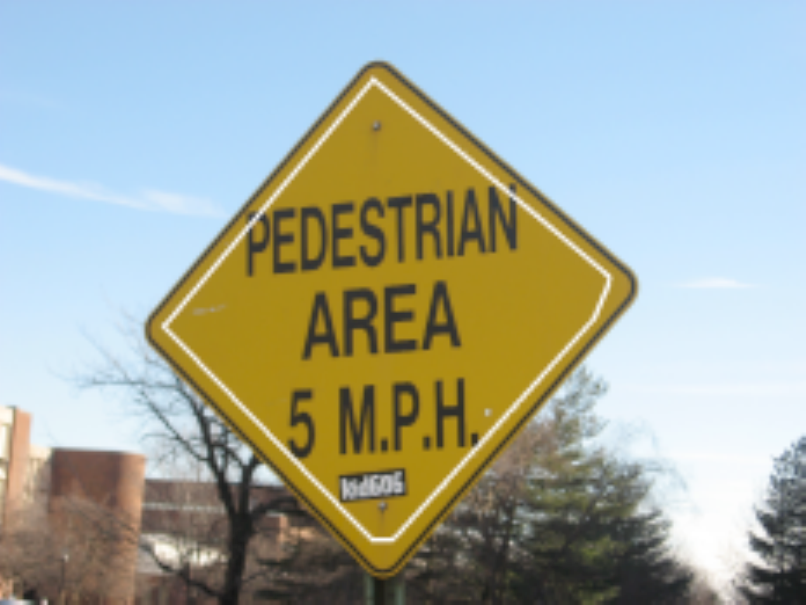,width=4cm}}
  \centerline{(b) }\medskip
\end{minipage}
\hfill
\begin{minipage}[b]{0.24\linewidth}
  \centering
 \centerline{\epsfig{figure=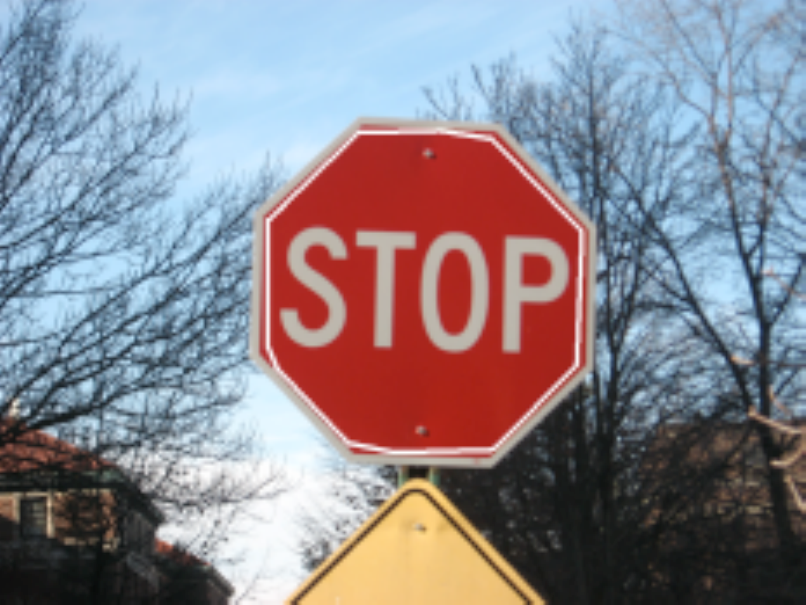,width=4cm}}
  \centerline{(c)  }\medskip
\end{minipage}
\hfill
\begin{minipage}[b]{0.24\linewidth}
  \centering
 \centerline{\epsfig{figure=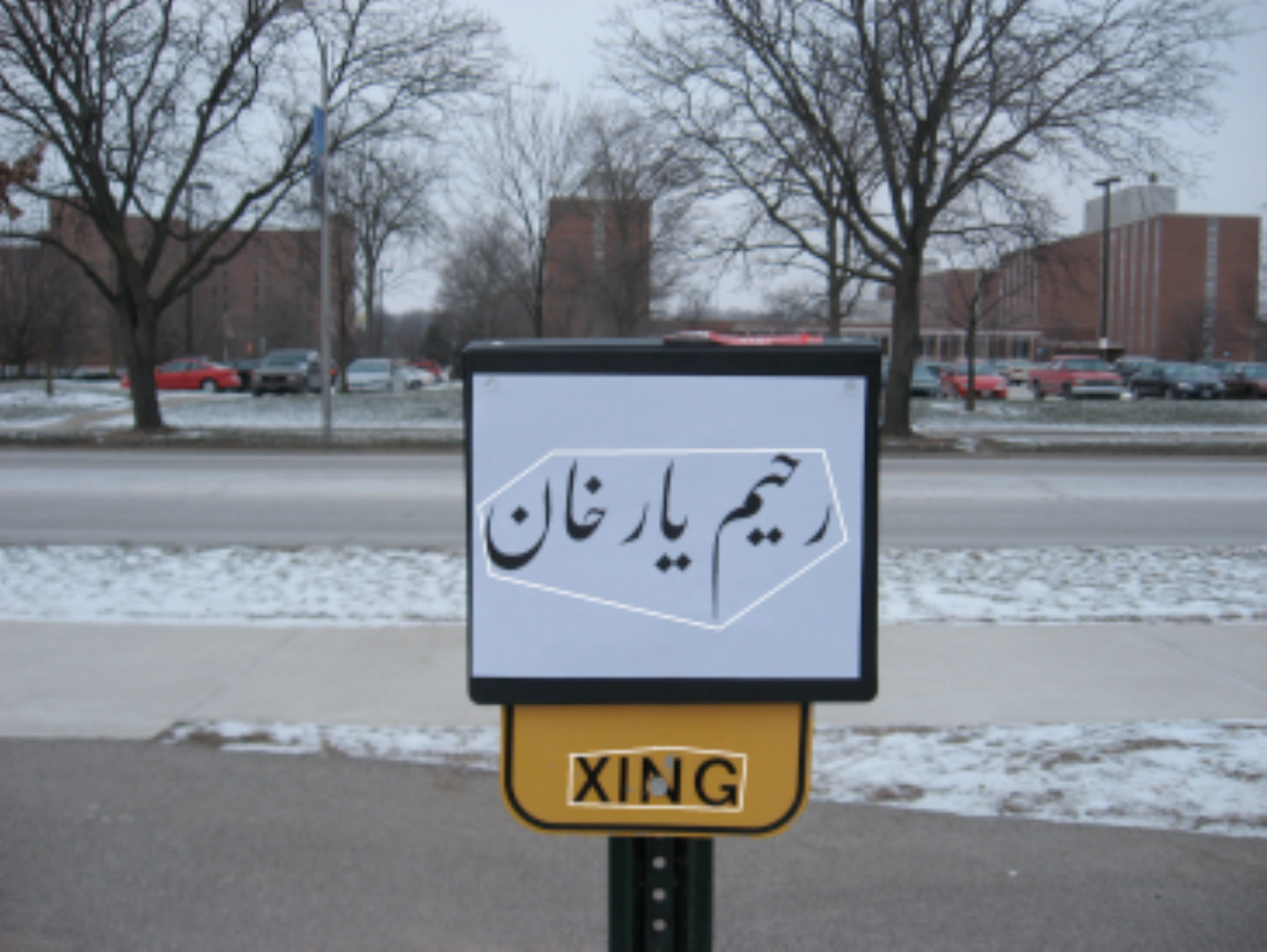,width=4cm}}
  \centerline{(d) }\medskip
\end{minipage}

\quad
\begin{minipage}[b]{0.24\linewidth}
  \centering
 \centerline{\epsfig{figure=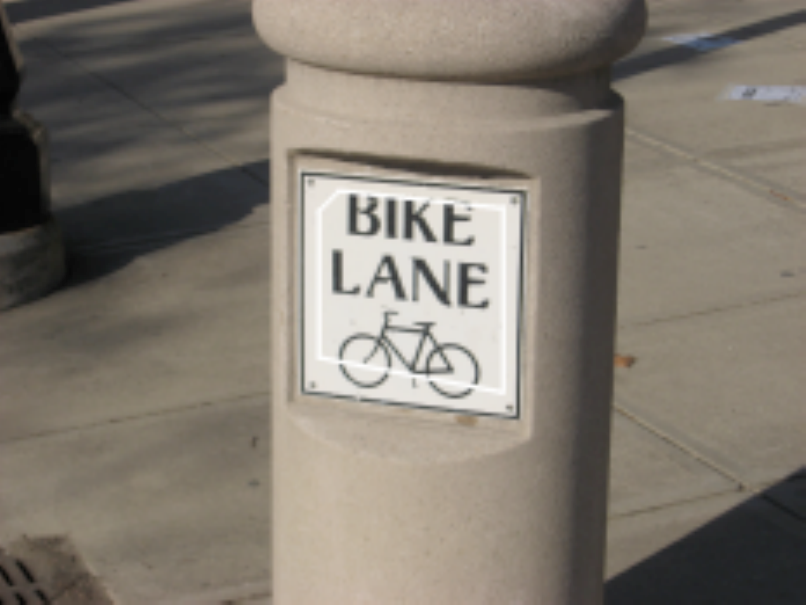,width=4cm}}
  \centerline{(e) }\medskip
\end{minipage}
\hfill
\begin{minipage}[b]{0.24\linewidth}
  \centering
 \centerline{\epsfig{figure=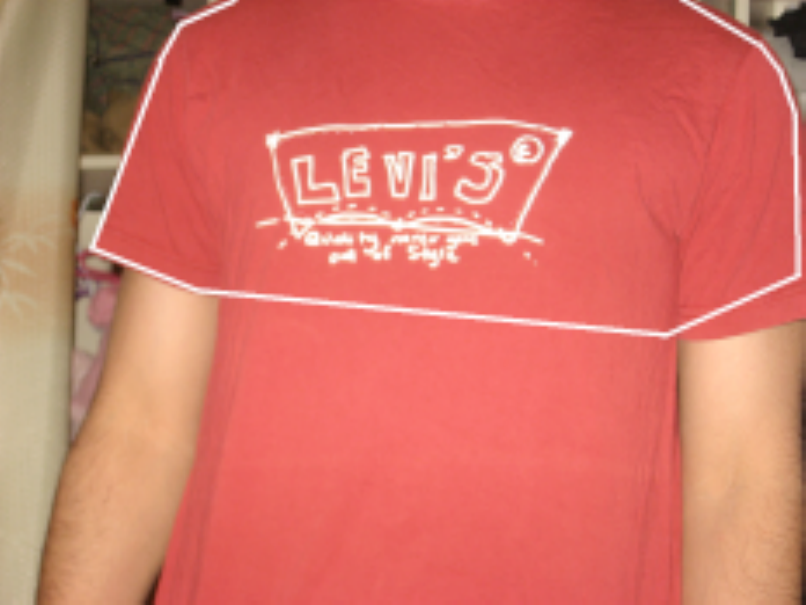,width=4cm}}
  \centerline{(f) }\medskip
\end{minipage}
\hfill
\begin{minipage}[b]{0.24\linewidth}
  \centering
 \centerline{\epsfig{figure=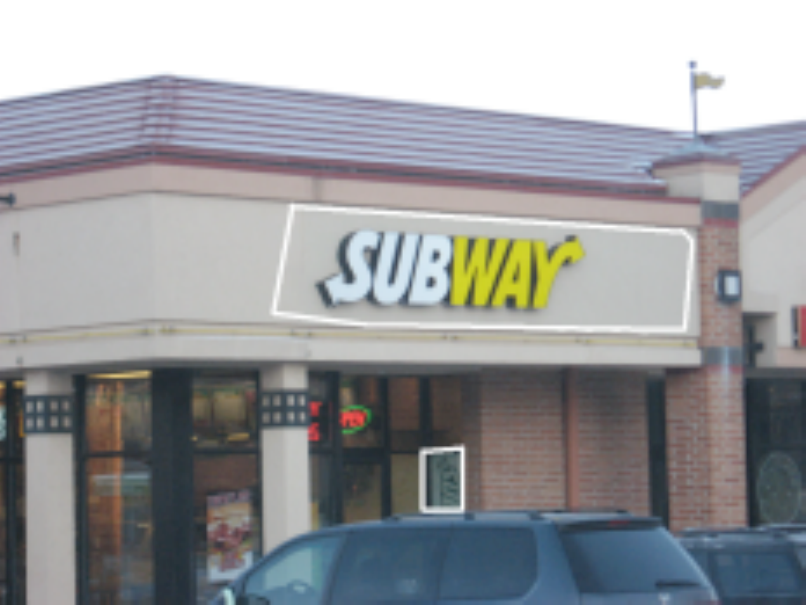,width=4cm}}
  \centerline{(g) }\medskip
\end{minipage}

\end{minipage}

\caption{A Few Samples of our Experimental Results.  
(a) Text of varying sizes and color (including graphics).
(b) Street sign in front of a smooth background (sky).
(c) Non-rectangular text area.
(d) Text written in English and Urdu both are successfully segmented.
(e) Street sign in front of a textured background (cement). 
(f) Text printed on an irregular surface.
(g) Shop display.
}
\label{fig:res}
\end{figure*}

\section{Experimental Results}
\label{sec:experiments}
We tested our method on a database of 65 (three megapixel) images of outdoor signs and shop displays. 
Ten of these images contained outdoor signs written in both English and Urdu. 
The rest (55 images) contained English signs only, but some included simple graphics as well.
All the text area was correctly segmented in 63 (i.e.,  97\%) of these 65 images. 
In four of these 63 images, some other areas were also segmented as well.
However, these areas all contain highly contrasting high level structures on a uniform background
which in many ways resemble text (for example, a capital "i" letter) but could be ruled out from a semantic point of view. 

\section{Conclusion}
\label{sec:conclusion}
We have presented a top-down hierarchical methods for finding text areas in natural images.
The key point of this method is that it begins by looking for text background areas
before testing for the presence of text inside the selected areas.
The method correctly segmented all the text in 97\% of the images in a small database of outdoor signs and shop displays.
 In future work, we will test the method on a larger database of natural images. 
To improve the results, we will use training to choose the optimal parameters 
for all the decisions we perform. 
We will also investigate the use of more sophisticated text presence test 
(e.g., edge based or connected component based)
 in order to improve the quality of the last step of our method.


\bibliographystyle{IEEEbib}
\ninept
\bibliography{strings}

\end{document}